\documentclass[runningheads]{llncs}

\usepackage{eccv}

\usepackage{eccvabbrv}

\usepackage{graphicx}
\graphicspath{ {figures/} }
\usepackage{booktabs}
\usepackage{bm}

\usepackage{multirow}
\usepackage{tabularx}
\usepackage[misc]{ifsym}

\usepackage{color}

\newcommand{\boldstart}[1]{\noindent\textbf{#1}}

\usepackage{etoolbox}
\newtoggle{withSupp}
\toggletrue{withSupp}

\usepackage[accsupp]{axessibility}  %

\usepackage{hyperref}

\newcommand{\method}{MVSplat\xspace}
\newcommand\rurl[1]{%
\href{https://#1}{\nolinkurl{#1}}%
}

\usepackage{orcidlink}

\begin{document}

\title{\texorpdfstring{MVSplat: Efficient 3D Gaussian Splatting \\
from Sparse Multi-View Images}{MVSplat: Efficient 3D Gaussian Splatting from Sparse Multi-View Images}} 

\titlerunning{MVSplat}

\author{Yuedong Chen\inst{1}\orcidlink{0000-0003-0943-1512}\textsuperscript{\Letter} \and
Haofei Xu\inst{2,3}\orcidlink{0000-0003-1313-3358} \and
Chuanxia Zheng\inst{4}\orcidlink{0000-0002-3584-9640} \and
Bohan Zhuang\inst{1}\orcidlink{0000-0002-0074-0303} \and \\
Marc Pollefeys\inst{2,5}\orcidlink{0000-0003-2448-2318} \and
Andreas Geiger\inst{3}\orcidlink{0000-0002-8151-3726} \and
Tat-Jen Cham\inst{6}\orcidlink{0000-0001-5264-2572} \and
Jianfei Cai\inst{1,6}\orcidlink{0000-0002-9444-3763}}

\authorrunning{Y.~Chen et al.}

\institute{$^1$Monash University ~~~ $^2$ETH Zurich ~~~ $^3$University of T\"ubingen, T\"ubingen AI Center \\
$^4$VGG, University of Oxford ~~~ $^5$Microsoft ~~~ $^6$Nanyang Technological University \\
\rurl{donydchen.github.io/mvsplat}
}

\maketitle

{
    \centering
    \captionsetup{type=figure}
    \includegraphics[width=\textwidth]{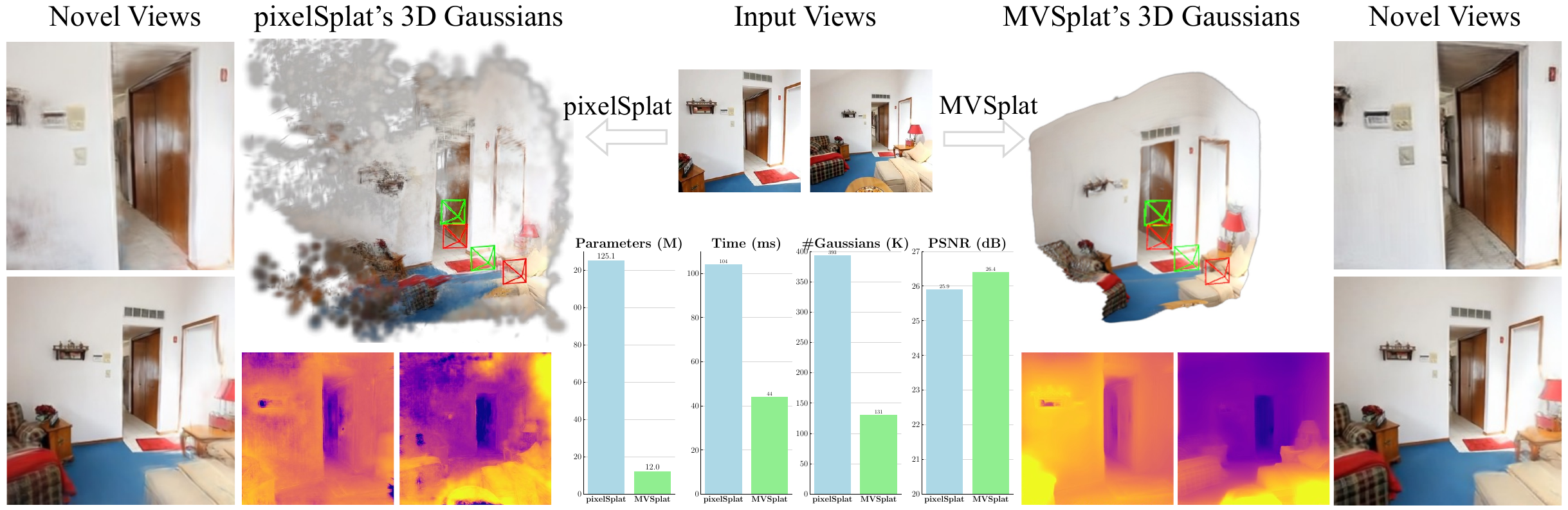}
    \captionof{figure}{
    Our MVSplat outperforms pixelSplat~\cite{charatan2023pixelsplat} in terms of both appearance and geometry quality with $10 \times$ fewer parameters and more than $2 \times$ faster inference speed.
    }
    \label{fig:teaser}
}

\begin{abstract}
  We introduce \method,
  an efficient model that,
  given sparse multi-view images as input,
  predicts clean feed-forward 3D Gaussians.
  To accurately localize the Gaussian centers,
  we build a cost volume representation via plane sweeping,
  where the cross-view feature similarities stored in the cost volume can provide valuable geometry cues to the estimation of depth.
  We also learn other Gaussian primitives' parameters
  jointly with the Gaussian centers while only relying on photometric supervision. 
  We demonstrate the importance of the cost volume representation in learning feed-forward Gaussians via extensive experimental evaluations.
  On the large-scale RealEstate10K and ACID benchmarks, \method achieves state-of-the-art performance with the fastest feed-forward inference speed (22~fps).
  More impressively, compared to the latest state-of-the-art method pixelSplat,
  \method uses $10\times $ fewer parameters and infers more than $2\times$ faster while providing higher appearance and geometry quality as well as better cross-dataset generalization.
\end{abstract}

\section{Introduction}
\label{sec:intro}

We consider the problem of 3D scene reconstruction and novel view synthesis from very sparse (\ie, as few as two) images in just one forward pass of a trained model.
While remarkable progress has been made using neural scene representations, \eg, Scene Representation Networks (SRN)~\cite{sitzmann2019scene}, Neural Radiance Fields (NeRF)~\cite{mildenhall2020nerf} and Light Filed Networks (LFN)~\cite{sitzmann2021light},
these methods are still not satisfactory for practical applications due to expensive per-scene optimization~\cite{niemeyer2022regnerf,truong2023sparf,wu2023reconfusion}, high memory cost~\cite{chen2021mvsnerf,johari2022geonerf,xu2024murf} and slow rendering speed~\cite{wang2021ibrnet,yu2021pixelnerf}.

Recently, 3D Gaussian Splatting (3DGS)~\cite{kerbl20233d} has emerged as an efficient and expressive 3D representation thanks to its fast rendering speed and high quality.
Using rasterization-based rendering, 3DGS inherently avoids the expensive volumetric sampling process of NeRF, %
leading to
highly efficient and high-quality 3D reconstruction and novel view synthesis.

Very recently, several feed-forward Gaussian Splatting methods have been proposed to explore 3D reconstruction from sparse view images, notably Splatter Image~\cite{szymanowicz2023splatter} and pixelSplat~\cite{charatan2023pixelsplat}.
Splatter Image regresses pixel-aligned Gaussian parameters using a standard image-to-image architecture,
which achieves promising results for single-view object-level 3D reconstruction.
However, reconstructing a 3D scene from a single image is inherently ill-posed and ambiguous, posing a significant challenge when applied to a more general and larger scene, which is the key focus of our paper.
pixelSplat~\cite{charatan2023pixelsplat} proposes to regress Gaussian parameters for the binocular reconstruction problem.
Specifically, it predicts a probabilistic depth distribution for each input view and then samples depths from that predicted distribution. 
Even though pixelSplat learns cross-view-aware features with an epipolar Transformer,
it is still challenging to predict a reliable probabilistic depth distribution solely from image features,
making pixelSplat's geometry reconstruction of comparably low quality and exhibiting noisy artifacts (see Fig.~\ref{fig:teaser} and Fig.~\ref{fig:point_cloud}).
For improved geometry reconstruction results, slow depth finetuning with an additional depth regularization loss is required.

To accurately localize the 3D Gaussian centers,
our solution is to build a cost volume representation via plane sweeping~\cite{collins1996space,yao2018mvsnet,xu2023unifying} in the 3D space.
Specifically,
the cost volume stores cross-view feature similarities for all potential depth candidates,
where the similarities can provide valuable geometry cues to the localization of 3D surfaces (\ie, high similarity more likely indicates a surface point).
With our cost volume representation,
the task is formulated as learning to perform \emph{feature matching} to identify the Gaussian centers, unlike the data-driven 3D \emph{regression} from image features in previous works~\cite{szymanowicz2023splatter,charatan2023pixelsplat}.
Such a formulation reduces the task's learning difficulty, enabling our method to achieve state-of-the-art performance with lightweight model size and fast speed.

We obtain 3D Gaussian centers by unprojecting the multi-view-consistent depths estimated by our constructed multi-view cost volumes with a 2D network.
Additionally, we also predict other Gaussian properties (covariance, opacity, and spherical harmonics coefficients), in parallel with the depths. 
This enables the rendering of novel view images using the predicted 3D Gaussians with the differentiable splatting operation~\cite{kerbl20233d}.
Our full model \method is trained end-to-end purely with the photometric loss between rendered and ground truth images.

On the large-scale RealEstate10K~\cite{DBLP:journals/tog/ZhouTFFS18} and ACID~\cite{liu2021infinite} benchmarks,
\method achieves state-of-the-art performance with the fastest feed-forward inference speed (22~fps).
More impressively,
compared to the state-of-the-art pixelSplat~\cite{charatan2023pixelsplat} (see Fig.~\ref{fig:teaser}), \method uses $10\times $ fewer parameters and infers more than $2\times$ faster while providing higher appearance and geometry quality as well as better cross-dataset generalization.
Furthermore, extensive ablation studies and analysis underscore the significance of our feature matching-based cost volume design in enabling highly efficient feed-forward 3D Gaussian Splatting models.

\section{Related Work}
\label{sec:related}

\boldstart{Sparse View Scene Reconstruction and Synthesis}.
The original NeRF and 3DGS methods are both designed for very dense views (\eg, 100) as inputs, which can be tedious to capture for real-world applications.
Recently, there have been growing interests~\cite{niemeyer2022regnerf,truong2023sparf,chen2021mvsnerf,chen2023explicit,xu2024murf,charatan2023pixelsplat,wu2023reconfusion,szymanowicz2024flash3d,gao2024cat3d,zhang2024gs} in scene reconstruction and synthesis from sparse input views (\eg, 2 or 3).
Existing sparse view methods can be broadly classified into two categories: per-scene optimization and cross-scene feed-forward inference methods.
Per-scene approaches mainly focus on designing effective regularization terms~\cite{niemeyer2022regnerf,truong2023sparf,yu2022monosdf,deng2022depth,wu2023reconfusion,fan2024instantsplat} to better constrain the optimization process.
However, they are inherently slow at inference time due to the expensive per-scene gradient back-propagation process.
In contrast, feed-forward models~\cite{yu2021pixelnerf,chen2021mvsnerf,chen2023explicit,xu2024murf,charatan2023pixelsplat,szymanowicz2023splatter,gao2024cat3d,zhang2024gs,szymanowicz2024flash3d} learn powerful priors from large-scale datasets, so that 3D reconstruction and view synthesis can be achieved via a single feed-forward inference by taking sparse views as inputs, which makes them significantly faster than those per-scene optimization methods.

\boldstart{Feed-Forward NeRF}.
Early approaches used NeRF~\cite{mildenhall2020nerf} for
\emph{objects}~\cite{Chibane_2021_CVPR,Henzler_2021_CVPR,johari2022geonerf,Liu_2022_CVPR,Reizenstein_2021_ICCV,wang2021ibrnet,yu2021pixelnerf} and \emph{scenes}~\cite{Chibane_2021_CVPR,du2023learning,chen2021mvsnerf,chen2023explicit,xu2024murf} 3D reconstrution.
pixelNeRF~\cite{yu2021pixelnerf} pioneered the paradigm of predicting pixel-aligned features from images for radiance field reconstruction.
The performance of feed-forward NeRF models progressively improved with the use of feature matching information~\cite{chen2021mvsnerf,chen2023explicit}, Transformers~\cite{sajjadi2022scene,du2023learning,miyato2023gta} and 3D volume representations~\cite{chen2021mvsnerf,xu2024murf}.
The state-of-the-art feed-forward NeRF model MuRF~\cite{xu2024murf} is based on a target view frustum volume and a (2+1)D CNN for radiance field reconstruction, where the 3D volume and CNN need to be constructed and inferred for every target view.
This makes MuRF expensive to train, with comparably slow rendering.
Most importantly, all existing feed-forward NeRF models suffer from the expensive per-pixel volume sampling in the rendering process.

\boldstart{Feed-Forward 3DGS}.
3D Gaussian Splatting~\cite{kerbl20233d,chen2024survey} avoids NeRF's expensive volume sampling via a rasterization-based splatting approach, where novel views can be rendered very efficiently from a set of 3D Gaussian primitives.
Very recently, a growing number of feed-forward 3DGS models~\cite{szymanowicz2023splatter,charatan2023pixelsplat,zheng2023gps,wewer2024latentsplat,zhang2024gs,chen2024lara,szymanowicz2024flash3d} have been proposed to solve the sparse-view-to-3D task. 
Among them, 
Splatter Image~\cite{szymanowicz2023splatter} proposes to regress pixel-aligned Gaussian parameters from a single view with a U-Net model. 
However, it mainly focuses on object-level reconstruction, while we target the more general scene-level reconstruction. 
Although its follow-up work Flash3D~\cite{szymanowicz2024flash3d} manages to extend to scene-level reconstruction, its performance on complex scenes is inherently non-satisfactory due to the limited information a single image can provide. In contrast, MVSplat is designed to effectively aggregate information from multi-view input.
More similar to our setting, 
pixelSplat~\cite{charatan2023pixelsplat} proposes to regress Gaussian parameters from two input views. It demonstrates the importance of cross-view-aware features, learned from the epipolar Transformer. 
It then directly uses these features to predict probabilistic depth distributions for sampling depths. However, the mapping from features to depth distributions is inherently ambiguous and unreliable, essentially leading to poor geometry reconstruction.
In contrast, we learn to predict depth from the feature matching information encoded within a cost volume, which makes it more geometry-aware and leads to a more lightweight model ($10\times$ fewer parameters and more than $2\times$ faster) and significantly better geometries.
Another related work GPS-Gaussian~\cite{zheng2023gps} proposes a feed-forward Gaussian model for the reconstruction of humans, instead of general scenes.
It relies on two rectified stereo images to estimate the disparity, while our method works for general unrectified multi-view posed images. During training, GPS-Gaussian requires ground truth depth for supervision, while our model is trained from RGB images alone.

\begin{figure}[t!]
    \centering
    \includegraphics[width=\textwidth]{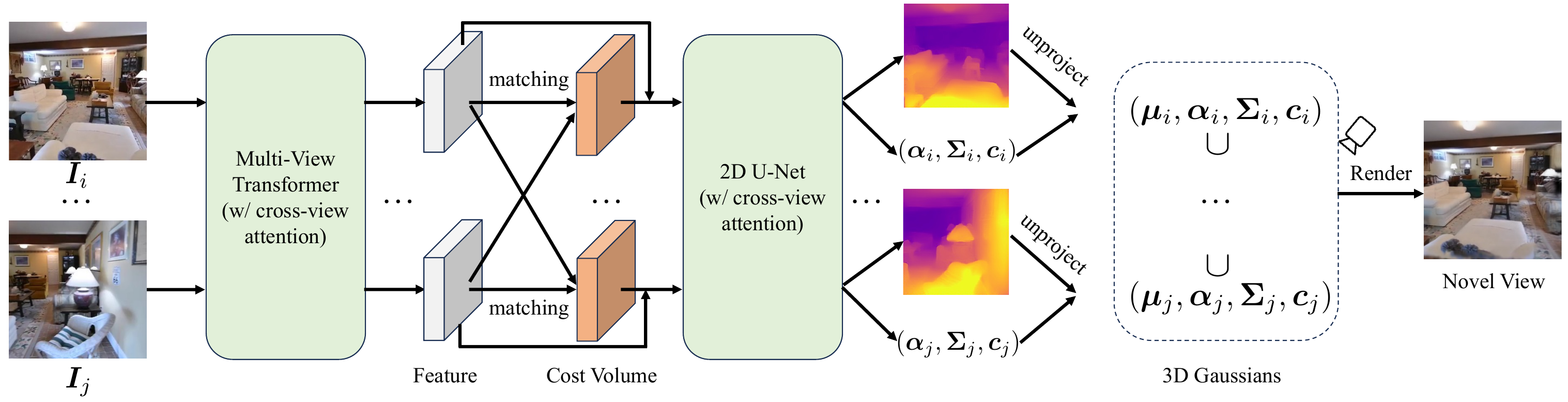}
    \caption{\textbf{Overview of MVSplat}.
    Given multiple posed images as input, \method first extracts multi-view image features with a Transformer.
    Then, the per-view cost volumes using plane sweeping are constructed.
    The Transformer features and cost volumes are concatenated together as input to a 2D U-Net (with cross-view attention) for cost volume refinement and predicting per-view depth maps.
    The per-view depth maps are unprojected to 3D and combined using a simple deterministic union operation as the 3D Gaussian centers.
    The opacity, covariance and color Gaussian parameters are predicted jointly with the depth maps. Finally, novel views are rendered from the predicted 3D Gaussians with the rasterization operation.
    }
    \label{fig:overview}
\end{figure}

\boldstart{Multi-View Stereo}. Multi-View Stereo (MVS) is a classic technique for reconstructing 3D scene structures from 2D images. Despite the conceptual similarity with the well-established MVS reconstruction pipeline~\cite{schonberger2016pixelwise,yao2018mvsnet,gu2020cascade}, our approach possesses unique advantages.
Unlike typical MVS methods involving \emph{separate} depth estimation and point cloud fusion stages, we exploit the unique properties of the 3DGS representation to infer 3D structure in a \emph{single} step, simply considering the union of the unprojected per-view depth predictions as the global 3D representation.
Besides, existing MVS networks~\cite{yao2018mvsnet,gu2020cascade,ding2022transmvsnet} are mostly trained with ground truth depth as supervision. 
In contrast, our model is fully differentiable and does \emph{not} require ground truth geometry supervision for training,
making it more scalable and suitable for in-the-wild scenarios.

\section{Method}
\label{sec:method}

We begin with $K$ sparse-view images $\mathcal{I}=\{{\bm I}^{i}\}_{i=1}^K$, (${\bm I}^i \in \mathbb{R}^{H \times W \times 3}$) and their corresponding camera projection matrices $\mathcal{P}=\{{\bm P}^i \}_{i=1}^K$, ${\bm P}^i=\mathbf{K}^i[\mathbf{R}^i|\mathbf{t}^i]$,
calculated via intrinsic $\mathbf{K}^i$, rotation $\mathbf{R}^i$ and translation $\mathbf{t}^i$ matrices.
Our goal is to learn a mapping $f_{\bm \theta}$ from images to 3D Gaussian parameters:
\begin{equation}
    f_{\bm \theta}: \{ ({\bm I}^{i}, {\bm P}^i) \}_{i=1}^K \mapsto \{(\bm{\mu}_j, \alpha_j, \bm{\Sigma}_j, \bm{c}_j )\}^{H \times W \times K}_{j=1},
\end{equation}
where we parameterize $f_{\bm \theta}$ as a feed-forward network and ${\bm \theta}$ are the learnable parameters optimized from a large-scale training dataset.
We predict the Gaussian parameters, including position $\bm{\mu}_j$, opacity $\alpha_j$, covariance $\bm{\Sigma}_j$ and color $\bm{c}_j$ (represented as spherical harmonics) in a pixel-aligned manner, and thus the total number of 3D Gaussians is $H \times W \times K$ for $K$ input images with shape $H \times W$.

To enable high-quality rendering and reconstruction, it is crucial to predict the position $\bm{\mu}_j$ precisely since it defines the center of the 3D Gaussian~\cite{kerbl20233d}.
In this paper, 
we present \method, a Gaussian-based feed-forward model for novel view synthesis.
Unlike pixelSplat~\cite{charatan2023pixelsplat} that predicts probabilistic depth,
we develop an efficient and high-performance multi-view depth estimation model that enables unprojecting predicted depth maps as the Gaussian centers,
in parallel with another branch for prediction of other Gaussian parameters ($\alpha_j$, $\bm{\Sigma}_j$ and $\bm{c}_j$).
Our full model, illustrated in~\cref{fig:overview}, is trained end-to-end using only a simple rendering loss for supervision. 
Next, we discuss the key components.

\subsection{Multi-View Depth Estimation}
\label{sec:mvdepth}

Our depth model is purely based on 2D convolutions and attentions, without any 3D convolutions used in many previous MVS~\cite{yao2018mvsnet,gu2020cascade,ding2022transmvsnet} and feed-forward NeRF~\cite{chen2021mvsnerf,xu2024murf} models.
This makes our model highly efficient.
Our depth model includes multi-view feature extraction, cost volume construction, cost volume refinement, depth estimation, and depth refinement, as introduced next.

\boldstart{Multi-view feature extraction}.
To construct the cost volumes, we first extract multi-view image features with a CNN and Transformer architecture~\cite{xu2022gmflow,xu2023unifying}.
Specifically, a shallow ResNet-like CNN is first used to extract $4 \times$ downsampled per-view image features.
Then, we use a multi-view Transformer with self- and cross-attention layers to exchange information between different views.
For better efficiency, we use Swin Transformer's local window attention~\cite{liu2021swin} in our Transformer architecture.
When more than two views are available, we perform cross-attention for each view with respect to all the other views, which has exactly the same learnable parameters as the 2-view scenario.
After this operation, we obtain \emph{cross-view aware} Transformer features $\{{\bm F}^{i}\}_{i=1}^K$ (${\bm F}^i \in \mathbb{R}^{\frac{H}{4} \times \frac{W}{4} \times C}$), where $C$ denotes the channel dimension.

\boldstart{Cost volume construction}. 
The key component of our model is the cost volume, which models cross-view feature matching information with respect to different depth candidates via the plane-sweep stereo approach~\cite{collins1996space,yao2018mvsnet,xu2023unifying}.
Note that we construct $K$ cost volumes for $K$ input views to predict $K$ depth maps.
Here, we take view $i$'s cost volume construction as an example.
Given the near and far depth ranges, we first uniformly sample $D$ depth candidates $\{ d_{m} \}_{m=1}^D$ in the inverse depth domain and then warp view $j$'s feature ${\bm F}^j$ to view $i$ with the camera projection matrices ${\bm P}^i$, ${\bm P}^j$ and each depth candidate $d_m$, to obtain $D$ warped features
\begin{equation}
\label{eq:warp}
    {\bm F}^{j \to i}_{d_m} = \mathcal{W}({\bm F}^j, {\bm P}^i, {\bm P}^j, d_m) \in \mathbb{R}^{\frac{H}{4} \times \frac{W}{4} \times C}, \quad m = 1, 2, \cdots, D,
\end{equation}
where $\mathcal{W}$ denotes the warping operation~\cite{xu2023unifying}. We then compute the dot product~\cite{xu2020aanet,xu2023unifying} between ${\bm F}^i$ and ${\bm F}^{j \to i}_{d_m}$ to obtain the correlation 
\begin{equation}
\label{eq:correlation}
    {\bm C}_{d_m}^{i} = \frac{{\bm F}^i \cdot {\bm F}^{j \to i}_{d_m}}{\sqrt{C}} \in \mathbb{R}^{\frac{H}{4} \times \frac{W}{4}}, \quad m = 1, 2, \cdots, D.
\end{equation}
When there are more than two views as inputs, we similarly warp another view's feature to view $i$ as in~\cref{eq:warp} and compute their correlations via~\cref{eq:correlation}. Finally, all the correlations are pixel-wise averaged, enabling the model to accept an arbitrary number of views as inputs.

Collecting all the correlations we obtain view $i$'s cost volume
\begin{equation}
\label{eq:cost_volume}
    {\bm C}^{i} = [{\bm C}_{d_1}^{i}, {\bm C}_{d_2}^{i}, \cdots, {\bm C}_{d_D}^{i}] \in \mathbb{R}^{\frac{H}{4} \times \frac{W}{4} \times D}.
\end{equation}
Overall, we obtain $K$ cost volumes $\{ {\bm C}^{i} \}_{i=1}^K$ for $K$ input views.

\boldstart{Cost volume refinement}. 
As the cost volume in~\cref{eq:cost_volume} can be ambiguous for texture-less regions,
we propose to further refine it with an additional lightweight 2D U-Net~\cite{ronneberger2015,rombach2022high}.
The U-Net takes the concatenation of Transformer features ${\bm F}^i$ and cost volume ${\bm C}^{i}$ as inputs, and outputs a residual $\Delta {\bm C}^{i} \in \mathbb{R}^{\frac{H}{4} \times \frac{W}{4} \times D}$ that is added to the initial cost volume ${\bm C}^{i}$. We obtain the refined cost volume as
\begin{equation}
    {\tilde{\bm C}}^{i} = {\bm C}^{i} + \Delta {\bm C}^{i} \in \mathbb{R}^{\frac{H}{4} \times \frac{W}{4} \times D}.
\end{equation}
To exchange information between cost volumes of different views, we inject three cross-view attention layers at the lowest resolution of the U-Net architecture.
This design ensures that the model can accept an arbitrary number of views as input since it computes the cross-attention for each view with respect to all the other views, where such an operation does \emph{not} depend on the number of views. 
The low-resolution cost volume ${\tilde{\bm C}}^{i}$ is finally upsampled to full resolution ${\hat{\bm C}}^{i} \in \mathbb{R}^{H \times W \times D}$ with a CNN-based upsampler.

\boldstart{Depth estimation}.
We use the $\texttt{softmax}$ operation to obtain per-view depth predictions.
Specifically, we first normalize the refined cost volume ${\hat{\bm C}}^{i}$ in the depth dimension and then perform a weighted average of all depth candidates ${\bm G} = [d_1, d_2, \cdots, d_D] \in \mathbb{R}^{D}$:
\begin{equation}
\label{eq:softmax_depth}
    {\bm V}^i = \mathrm{softmax} (\hat{\bm C}^i) {\bm G} \in \mathbb{R}^{H \times W}.
\end{equation}

\boldstart{Depth refinement}.
To further improve the performance, we introduce an additional depth refinement step to enhance the quality of the predicted depth.
The refinement is performed with a very lightweight 2D U-Net, which takes multi-view images, features, and current depth predictions as input, and outputs per-view residual depths.
The residual depths are then added to the current depth predictions as the final depth outputs.
Similar to the U-Net used in the above cost volume refinement, we also introduce cross-view attention layers in the lowest resolution to exchange information across views. 
More implementation details are presented in the supplementary material 
\iftoggle{withSupp}{\cref{sec:app_implementation}}{Appendix~C}.

\subsection{Gaussian Parameters Prediction} 
\label{sec:3dgs_dec}
\boldstart{Gaussian centers $\bm \mu$}.
After obtaining the multi-view depth predictions, we directly unproject them to 3D point clouds using the camera parameters.
We transform the per-view point cloud into an aligned world coordinate system and directly combine them as the centers of the 3D Gaussians.

\boldstart{Opacity $\alpha$}.
From the matching distribution obtained via the $\mathrm{softmax} (\hat{\bm C}^i)$ operation in~\cref{eq:softmax_depth}, we can also obtain the matching confidence as the maximum value of the softmax output.
Such matching confidence shares a similar physical meaning with the opacity (points with higher matching confidence are more likely to be on the surface), and thus we use two convolution layers to predict the opacity from the matching confidence input.

\boldstart{Covariance $\bm \Sigma$ and color $\bm c$}.
We predict these parameters using two convolution layers that take as inputs the concatenated image features, refined cost volume, and original multi-view images.
Similar to other 3DGS approaches~\cite{kerbl20233d,charatan2023pixelsplat}, the covariance matrix $\bm{\Sigma}=
R(\theta)^\mathsf{T} \operatorname{diag}(s) R(\theta)$ is composed of a scaling matrix
$s$
and a rotation matrix
$R(\theta)$
represented via quaternions, and the color $\bm {c}$ is calculated from the predicted spherical harmonic coefficients.

\subsection{Training Loss}

Our model predicts a set of 3D Gaussian parameters $\{(\bm{\mu}_j, \alpha_j, \bm{\Sigma}_j, \bm{c}_j)\}^{H \times W \times K}_{j=1}$,
which are then used for rendering images at novel viewpoints.
Our full model is trained with ground truth target RGB images as supervision.
The training loss is calculated as a linear combination of $\ell_2$ and LPIPS~\cite{zhang2018unreasonable} losses, with loss weights of 1 and 0.05, respectively.

\section{Experiments}
\label{sec:exp}

\subsection{Settings}

\boldstart{Datasets.} 
We assess our model on the large-scale RealEstate10K~\cite{DBLP:journals/tog/ZhouTFFS18} and ACID~\cite{liu2021infinite} datasets.
RealEstate10K contains real estate videos downloaded from YouTube, which are split into 67,477 {training} scenes and 7,289 {testing} scenes,
while ACID contains nature scenes captured by aerial drones,
which are split into 11,075 {training} scenes and 1,972 {testing} scenes.
Both datasets provide estimated camera intrinsic and extrinsic parameters for each frame.
Following pixelSpalt~\cite{charatan2023pixelsplat},
we evaluate all methods on three target novel viewpoints for each {test} scene.
Furthermore, to further assess the cross-dataset generalization ability,
we also directly evaluate all models on the multi-view DTU~\cite{jensen2014large} dataset,
which contains object-centric scenes with camera poses. 
On the DTU dataset, we report results on 16 validation scenes, with 4 novel views for each scene. 

\boldstart{Metrics.} 
For quantitative results, we report the standard image quality metrics,
including pixel-level PSNR, patch-level SSIM~\cite{wang2004image}, and feature-level LPIPS~\cite{zhang2018unreasonable}.
The inference time and model parameters are also reported to enable thorough comparisons of speed and accuracy trade-offs.
For a fair comparison, all experiments are conducted on $256\times256$ resolutions following existing models~\cite{charatan2023pixelsplat,szymanowicz2023splatter}.

\boldstart{Implementation details.}
\method is implemented with PyTorch, along with an off-the-shelf 3DGS render implemented in CUDA.
Our multi-view Transformer contains 6 stacked self- and cross-attention layers.
We sample 128 depth candidates when constructing the cost volumes in all the experiments.
All models are trained on a single A100 GPU for 300,000 iterations with the Adam~\cite{kingma2014adam} optimizer.
More details are provided in the supplementary material \iftoggle{withSupp}{\cref{sec:app_implementation}}{Appendix~C}.
Code and models are available at \url{https://github.com/donydchen/mvsplat}.

\subsection{Main Results} \label{sec:exp_comparisons}

\begin{table*}[t]
    \begin{center}
\footnotesize
    \setlength{\tabcolsep}{2.5pt} %
    \begin{tabular}{@{}l c c ccc c ccc@{}}
    \toprule
    \multirow{2}{*}[-2pt]{\textbf{Method}} & \multirow{2}{*}[-2pt]{\begin{tabular}[x]{@{}c@{}}\textbf{Time}\\(s) \end{tabular}} & \multirow{2}{*}[-2pt]{\begin{tabular}[x]{@{}c@{}}\textbf{Param}\\(M) \end{tabular}} & \multicolumn{3}{c}{\textbf{RealEstate10K~\cite{DBLP:journals/tog/ZhouTFFS18}}} && \multicolumn{3}{c}{\textbf{ACID~\cite{liu2021infinite}}} \\
    \addlinespace[-12pt] \\
    \cmidrule{4-6} \cmidrule{8-10} 
    \addlinespace[-12pt] \\
    & & & PSNR$\uparrow$ & SSIM$\uparrow$ & LPIPS$\downarrow$ && PSNR$\uparrow$ & SSIM$\uparrow$ & LPIPS$\downarrow$ \\
    \midrule

    pixelNeRF~\cite{yu2021pixelnerf} & 5.299 & 28.2 & 20.43 & 0.589 & 0.550 && 20.97 & 0.547 & 0.533 \\
    GPNR~\cite{suhail2022generalizable} & 13.340 & 9.6 & 24.11 & 0.793 & 0.255 && 25.28 & 0.764 & 0.332 \\
    AttnRend~\cite{du2023learning} & 1.325 & 125.1  & 24.78 & 0.820 & 0.213 && 26.88 & 0.799 & 0.218 \\
    MuRF~\cite{xu2024murf} & 0.186 & \textbf{5.3}  & 26.10 & 0.858 & 0.143 && 28.09 & 0.841 & 0.155 \\
    \midrule
    pixelSplat~\cite{charatan2023pixelsplat} & 0.104 & 125.4  & 25.89 & 0.858 & 0.142 && 28.14 & 0.839 & 0.150 \\
    \textbf{MVSplat} & \textbf{0.044} & 12.0 & \textbf{26.39} & \textbf{0.869} & \textbf{0.128} && \textbf{28.25} & \textbf{0.843} & \textbf{0.144}  \\
    \bottomrule
    \end{tabular}
    \end{center}
    \caption{\textbf{Comparisons with the state of the art}.
    Running time includes both encoder and render, note that 3DGS-based methods (pixelSplat and MVSplat) render dramatically faster ($\sim 500$FPS for the render).
    Performances are averaged over thousands of test scenes in each dataset.
    For each scene, the model takes two views as input and renders three novel views for evaluation.
    MVSplat performs the best in terms of all visual metrics and runs the fastest with a lightweight model size.
    }
    \label{tab:sota_compare}
\end{table*}

\boldstart{Baselines.}
We compare \method with several representative feed-forward methods that focus on scene-level novel view synthesis from sparse views,
including \textbf{i)} Light Field Network-based GPNR~\cite{suhail2022generalizable} and AttnRend~\cite{du2023learning}, 
\textbf{ii)} NeRF-based pixelNeRF~\cite{yu2021pixelnerf} and MuRF~\cite{xu2024murf},
\textbf{iii)} the latest state-of-the-art 3DGS-based model pixelSplat~\cite{charatan2023pixelsplat}.
We conduct thorough comparisons with the latter, being the most closely related to our method.

\boldstart{Assessing image quality.}
We report results on the RealEstate10K~\cite{DBLP:journals/tog/ZhouTFFS18} and ACID~\cite{liu2021infinite} benchmarks in~\cref{tab:sota_compare}.
\method surpasses all previous state-of-the-art models in terms of all metrics on visual quality,
with more obvious improvements in the LPIPS metric,
which is better aligned with human perception.
This includes pixelNeRF~\cite{yu2021pixelnerf}, GPNR~\cite{suhail2022generalizable}, AttnRend~\cite{du2023learning} and pixelSplat~\cite{charatan2023pixelsplat},
with results taken directly from the pixelSplat~\cite{charatan2023pixelsplat} paper,
and the recent state-of-the-art NeRF-based method MuRF~\cite{xu2024murf}, for which we 
re-train and evaluate its performance using the officially released code.

\begin{figure}[t!]
    \centering
    \includegraphics[width=\textwidth]{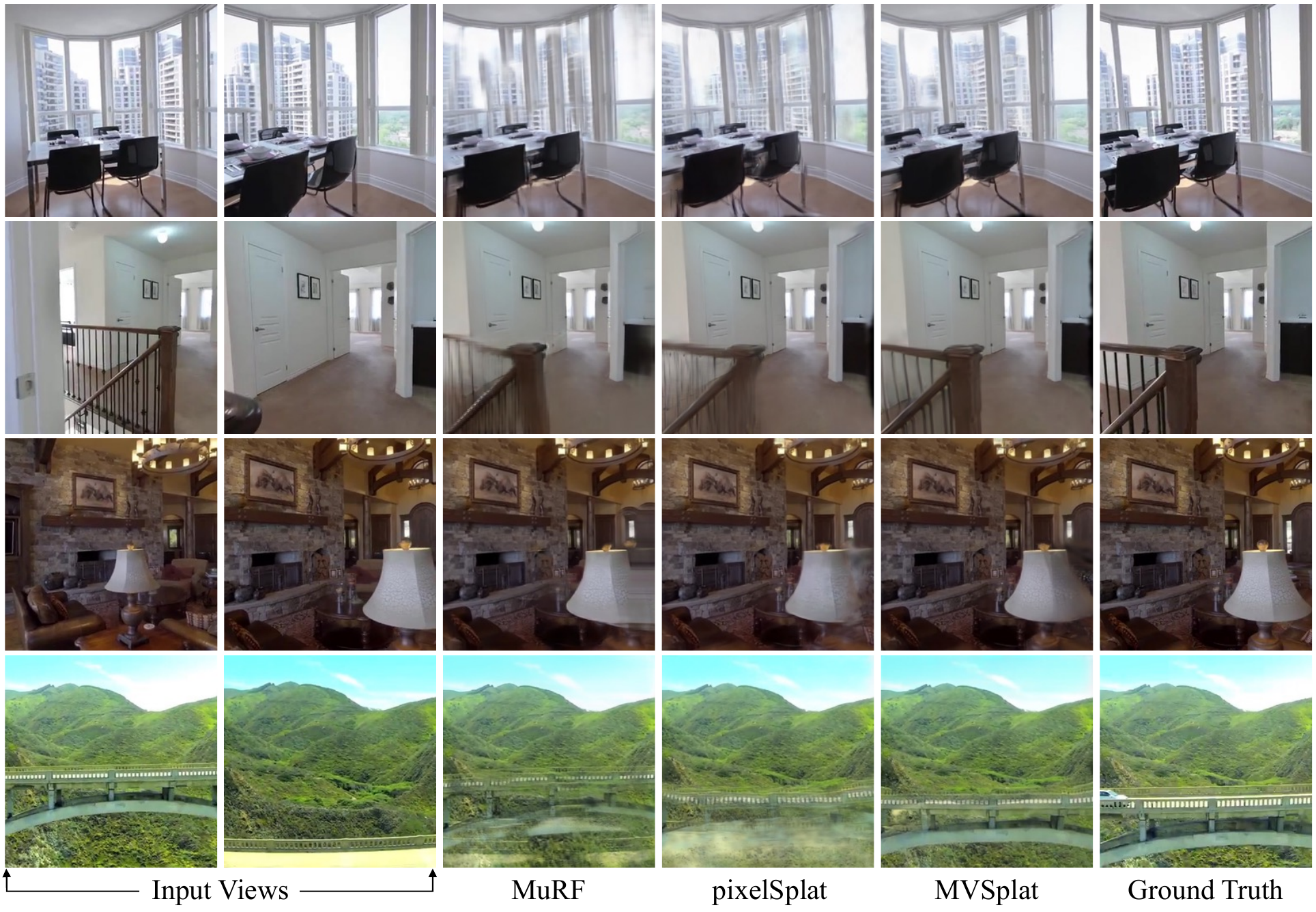}
    \begin{picture}(0,0)
    \put(43,15){\scriptsize \cite{charatan2023pixelsplat}}
    \put(-21,15){\scriptsize \cite{xu2024murf}}
    \end{picture}
    \caption{\textbf{Comparisons with the state of the art}. The first three rows are from RealEstate10K (indoor scenes), while the last one is from ACID (outdoor scenes).
    Models are trained with a collection of training scenes from each indicated dataset, and tested on novel scenes from the same dataset. 
    \method surpasses all other competitive models in rendering challenging regions due to the effectiveness of our cost volume-based geometry representation.}
    \label{fig:sota_comparisons}
\end{figure}

The qualitative comparisons of the top three best models are visualized in~\cref{fig:sota_comparisons}.
\method achieves the highest quality on novel view results even under challenging conditions,
such as these regions with repeated patterns (``window frames'' in 1st row),
or these present in only one of the input views (``stair handrail'' and ``lampshade'' in 2nd and 3rd rows),
or when depicting large-scale outdoor objects captured from distant viewpoints (``bridge'' in 4th row).
The baseline methods exhibit obvious artifacts for these regions,
while \method shows no such artifacts due to our cost volume-based geometry representation.
More evidence and detailed analysis regarding how \method effectively infers the geometry structures are presented in Sec.~\ref{sec:exp_abl}.

\boldstart{Assessing model efficiency.}
As reported in Tab.~\ref{tab:sota_compare}, apart from attaining superior image quality,
\method also shows the fastest inference time among all the compared models,
accompanied by a lightweight model size,
demonstrating its efficiency and practical utility.
It is noteworthy that the reported time encompasses both image encoding and rendering stages.
For an in-depth time comparison with pixelSplat~\cite{charatan2023pixelsplat},
our encoder runs at 0.043s, which is more than $2\times$ faster than pixelSplat (0.102s).
Besides, pixelSplat predicts 3 Gaussians per-pixel,
while our \method predicts 1 single Gaussian, which also contributes to our faster rendering speed (0.0015s \vs 0.0025s) due to the threefold reduction in the number of Gaussians.
More importantly, equipped with the cost volume-based encoder, our \method enables fast feed-forward inference of 3D Gaussians with a much light-weight design, resulting in $10 \times$ fewer parameters and more than $2\times$ faster speed compared to pixelSplat~\cite{charatan2023pixelsplat}.

\begin{figure}[t!]
    \centering
    \includegraphics[width=\textwidth]{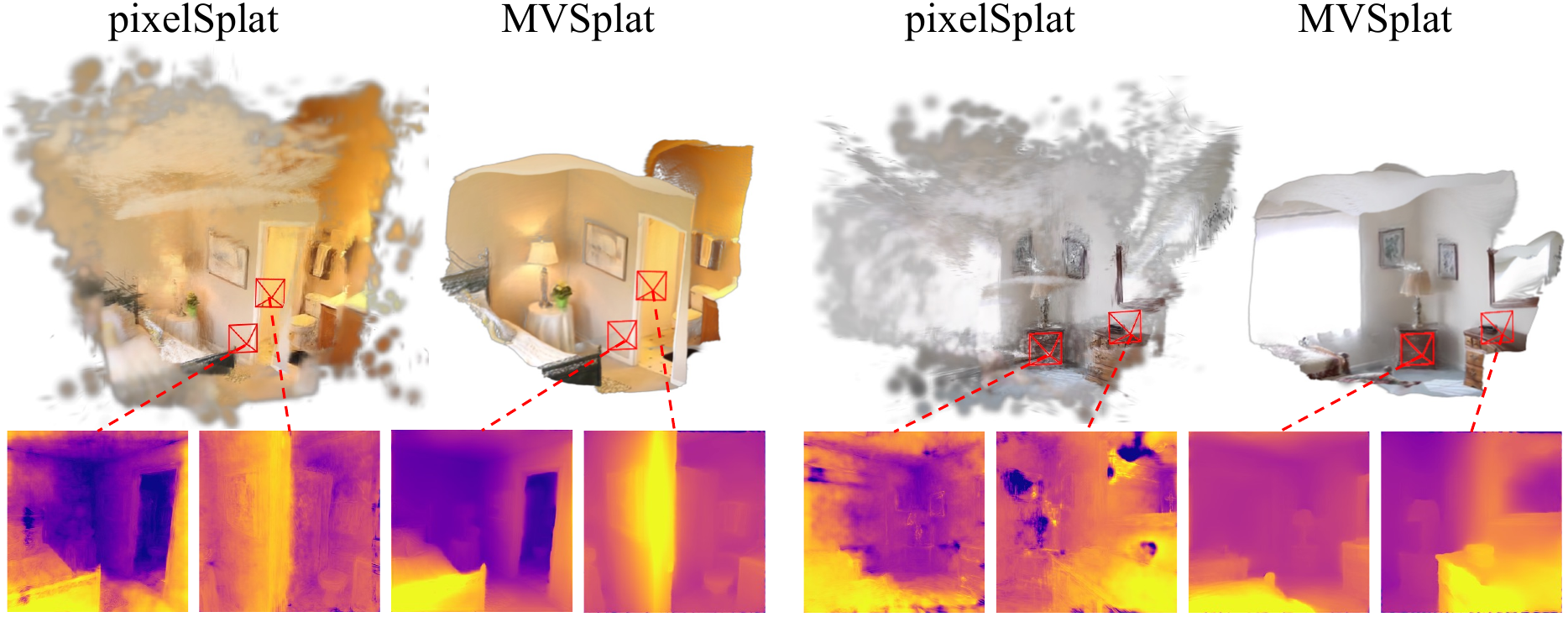}
    \begin{picture}(0,0)
    \put(-113,141){\footnotesize \cite{charatan2023pixelsplat}}
    \put(63,141){\footnotesize \cite{charatan2023pixelsplat}}
    \end{picture}
    \caption{\textbf{Comparisons of 3D Gaussians (top) and depth maps (bottom)}. 
    We compare the reconstructed geometry quality by visualizing zoom-out views of 3D Gaussians predicted by pixelSplat and our \method, along with the predicted depth maps of two reference views. Extra fine-tuning is \emph{not} performed on either model. Unlike pixelSplat that contains obvious floating artifacts, our \method produces much higher quality 3D Gaussians and smoother depth, demonstrating the effectiveness of our cost volume-based 3D representation.
    }
    \label{fig:point_cloud}
\end{figure}

\boldstart{Assessing geometry reconstruction.}
\method also produces significantly higher-quality 3D Gaussian primitives compared to the latest state-of-the-art pixelSplat~\cite{charatan2023pixelsplat}, as demonstrated in \cref{fig:point_cloud}.
pixelSplat requires an extra 50,000 steps to fine-tune the Gaussians with an additional depth regularization to achieve reasonable geometry reconstruction results.
Our \method instead generates high-quality geometries by \emph{training solely with photometric supervision}. 
\cref{fig:point_cloud} demonstrates the feed-forward geometry reconstruction results of \method, without any extra fine-tuning.
Notably, although pixelSplat showcases reasonably rendered 2D images, its underlying 3D structure contains a large amount of floating Gaussians.
In contrast, our \method reconstructs much higher-quality 3D Gaussians, demonstrating the effectiveness of our cost volume representation in obtaining high-quality geometry structures.
To facilitate a clearer understanding of the difference between the two methods, we invite the reader to view the corresponding ``.ply'' files of the exported 3D Gaussians on our project page.

\begin{figure}[t!]
    \centering
    \includegraphics[width=\textwidth]{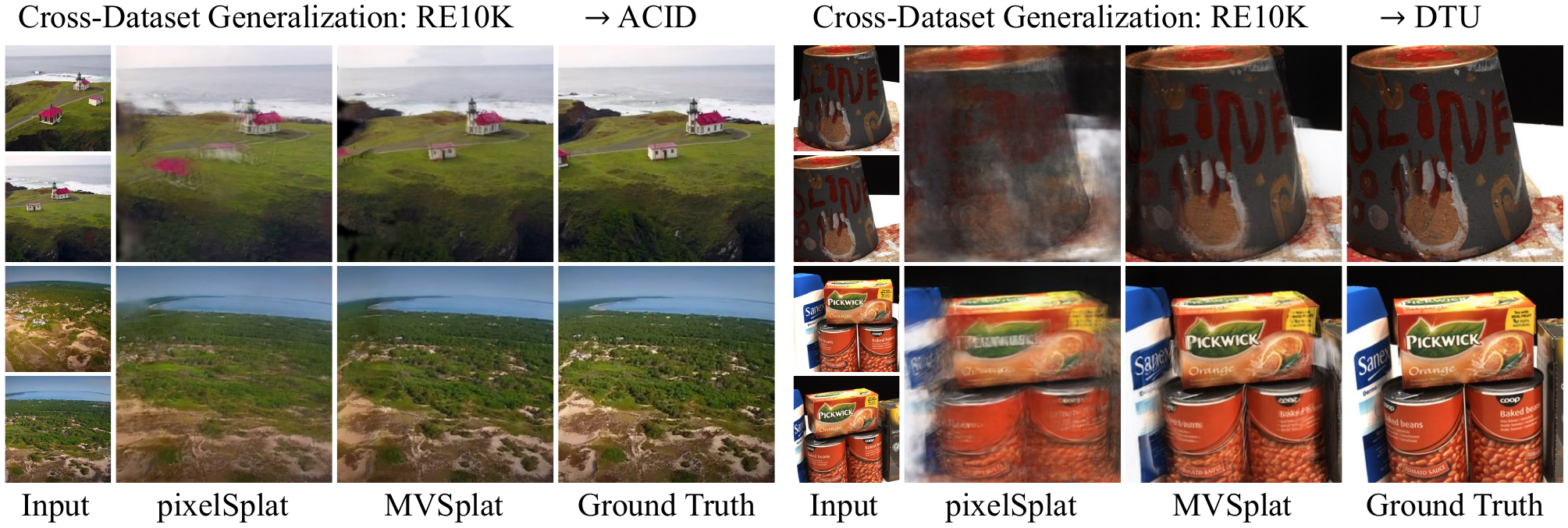}
    \begin{picture}(0,0)
    \put(-61,123){\scriptsize \cite{DBLP:journals/tog/ZhouTFFS18}}
    \put(-20.5,123){\scriptsize \cite{liu2021infinite}}
    \put(115,123){\scriptsize \cite{DBLP:journals/tog/ZhouTFFS18}}
    \put(153,123){\scriptsize \cite{jensen2014large}}  
    \put(-111.5,14){\scriptsize \cite{charatan2023pixelsplat}}
    \put(63,14){\scriptsize \cite{charatan2023pixelsplat}}
    \end{picture}
    \caption{\textbf{Cross-dataset generalization}. Models trained on the source dataset RealEstate10K (indoor scenes) are used to conduct zero-shot test on scenes from target datasets ACID (outdoor scenes) and DTU (object-centric scenes), without any fine-tuning. pixelSplat tends to render blurry images with obvious artifacts since feature distributions in the target datasets differ from the one in the source, while our \method renders competitive outputs thanks to the feature-invariant cost volume based design.}
    \label{fig:re10k_generalization}
\end{figure}

\begin{table}[t]
    \begin{center}
    \begin{tabular}{lccccccccccccccc}
    \toprule

    \multirow{2}{*}[-2pt]{Training data} & \multirow{2}{*}[-2pt]{Method} & \multicolumn{3}{c}{ACID~\cite{liu2021infinite}} & \multicolumn{3}{c}{DTU~\cite{jensen2014large}} \\
    \addlinespace[-12pt] \\
    \cmidrule(lr){3-5} \cmidrule(lr){6-8} 
    \addlinespace[-12pt] \\
    & & PSNR$\uparrow$ & SSIM$\uparrow$ & LPIPS$\downarrow$ & PSNR$\uparrow$ & SSIM$\uparrow$ & LPIPS$\downarrow$ \\
    
    \midrule

    \multirow{2}{*}[-2pt]{RealEstate10K~\cite{DBLP:journals/tog/ZhouTFFS18}} & pixelSplat~\cite{charatan2023pixelsplat} &  27.64 & 0.830 & 0.160 & 12.89 & 0.382 & 0.560 \\
                                         & \method & \textbf{28.15} & \textbf{0.841} & \textbf{0.147}  & \textbf{13.94} & \textbf{0.473} & \textbf{0.385}  \\

    \bottomrule
    \end{tabular}
    \end{center}
    \caption{\textbf{Cross-dataset generalization}. Models trained on RE10K (indoor scenes) are directly used to test on scenes from ACID (outdoor scenes) and DTU (object-centric scenes), without any further fine-tuning. Our \method generalizes better than pixelSplat, where the improvement is more significant when the gap between source and target datasets is larger (RE10K to DTU).  
    It is also worth noting that our zero-shot generalization results on ACID even slightly surpass pixelSplat's ACID trained model (PSNR: 28.14, SSIM: 0.843, LPIPS: 0.144) reported in Tab.~\ref{tab:sota_compare}. 
    }
    \label{tab:generalization}

\end{table}

\boldstart{Assessing cross-dataset generalization.}
\method is inherently superior in generalizing to \emph{out-of-distribution} novel scenes,
primarily due to the fact that the cost volume captures the \emph{relative similarity} between features, which remains \emph{invariant} compared to the absolute scale of features.
To demonstrate this advantage, we conduct two cross-dataset evaluations. 
Specifically, we choose models trained solely on the RealEstate10K (indoor scenes), and directly test them on ACID (outdoor scenes) and DTU (object-centric scenes).
As evident in~\cref{fig:re10k_generalization}, \method renders competitive novel views, despite scenes of the targeted datasets containing significantly different camera distributions and image appearance from those of the source dataset.
In contrast, views rendered by pixelSplat degrade dramatically; the main reason is likely that pixelSplat relies on 
purely feature aggregations that are tied to the absolute scale of feature values,
hindering its performance when it receives different image features from other datasets.
Quantitative results reported in Tab.~\ref{tab:generalization} further uphold this observation.
Note that the \method significantly outperforms pixelSplat in terms of LPIPS, and the gain is larger when the domain gap between source and target datasets becomes larger.
More surprisingly, our cross-dataset generalization results on ACID even slightly surpass the pixelSplat model that is specifically trained on ACID (see Tab.~\ref{tab:sota_compare}).
We attribute such results to the larger scale of the RealEstate10K training set ($\sim 7 \times$ larger than ACID) and our superior generalization ability.
This also suggests the potential of our method for training on more diverse and larger-scale datasets in the future.

\boldstart{Assessing more-view quality.}
\method is designed to be agnostic to the number of input views, so that it can benefit from more input views if they are available in the testing phase, regardless of how many input views are used in training.
We verify this by testing on DTU with 3 context views, using the model trained on the 2-view RealEstate10K dataset.
Our results are PSNR: 14.30, SSIM: 0.508, LPIPS: 0.371, and pixelSplat's are PSNR: 12.52, SSIM: 0.367, LPIPS: 0.585. 
Compared to the 2-view results (Tab.~\ref{tab:generalization}), \method achieves better performance with more input views. 
However, pixelSplat performs slightly worse when using more views, even though we have made our best effort to extend its released 2-views-only model to support more-view testing.
This suggests that the feature distribution of more views might be different from the two views used to train pixelSplat.
This discrepancy is attributed to the reliance of pixelSplat on pure feature aggregation, which lacks robustness to changes in feature distribution. 
This limitation is analogous to the reason why pixelSplat performs inferior in cross-dataset generalization tests discussed earlier.

\begin{figure}[t!]
    \centering
    \includegraphics[width=\textwidth]{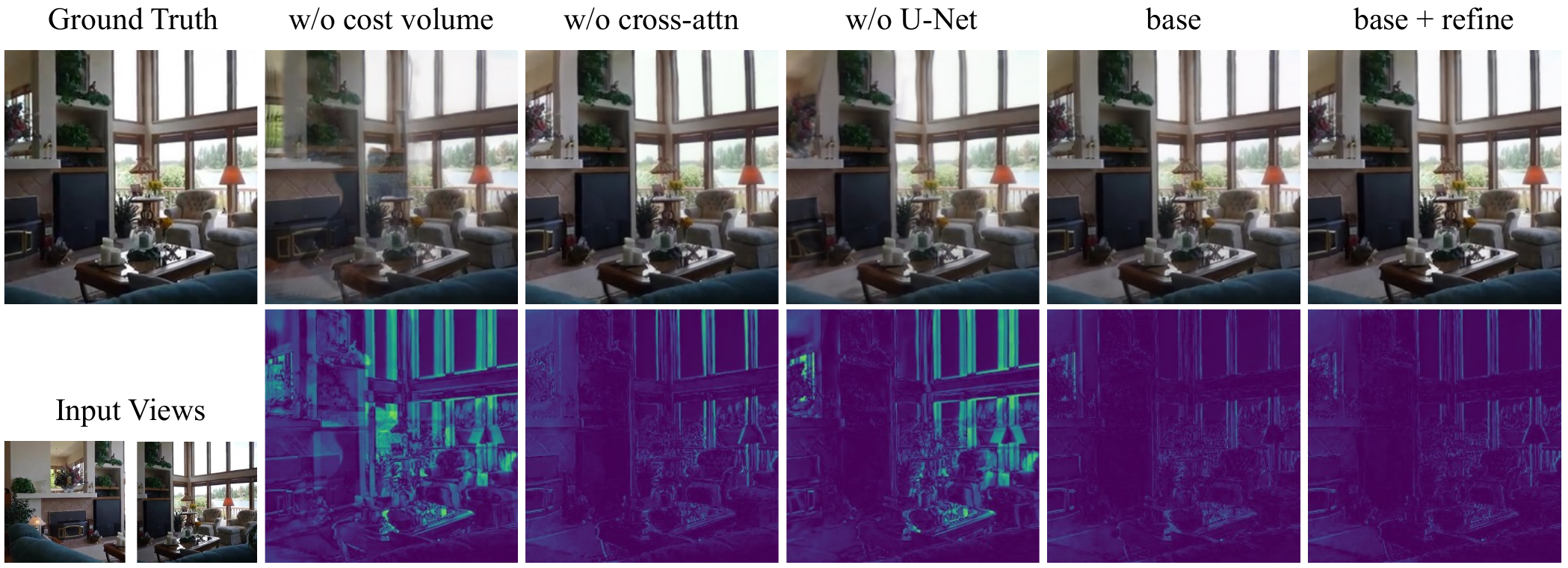}
    \caption{\textbf{Ablations on RealEstate10K}.
    Colored \emph{error maps} obtained by calculating the differences between the rendered images and the ground truth are attached for better comparison.
    All models are based on the ``base'' model, w/o depth refinement module.
    ``w/o cross-attn'' is short for ``w/o cross-view attention''.
    Compared to ``base'', ``base+refine'' slightly reduces errors, ``w/o cost volume'' leads to the largest drop, ``w/o U-Net'' harms contents on the right that only exist in one input view,
    while ``w/o cross-attn'' increases the overall error intensity.}
    \label{fig:re10k_ablation}
    
\end{figure}

\begin{table}[t]
    \begin{center}
    \begin{tabular}{llcccccccccccccc}
    \toprule
    
    Setup  & PSNR$\uparrow$ & SSIM$\uparrow$ & LPIPS$\downarrow$  \\
    
    \midrule

    base + refine & \textbf{26.39} & \textbf{0.869} & \textbf{0.128} \\
    \midrule

    base & 26.12 & 0.864 & 0.133 \\
    w/o cost volume & 22.83  & 0.753  & 0.197 \\
    w/o cross-view attention & 25.19 & 0.852 & 0.152 \\
    w/o U-Net & 25.45 & 0.847 & 0.150 \\
    
    \bottomrule
    \end{tabular}
    \end{center}
    \caption{\textbf{Ablations on RealEstate10K}.
    The ``base + refine'' is our final model, where ``refine'' refers to the ``depth refinement'' detailed in Sec.~\ref{sec:mvdepth}.
    All other ablations are conducted on the ``base'' model w/o depth refinement.
    The cost volume module plays an indispensable role in \method.
    Results of all models are obtained from the final converged step (300K in our experiments), except the one ``w/o cross-view attention'', which suffers from over-fitting
    (details are shown in the supplementary material \iftoggle{withSupp}{\cref{fig:supp_val_logs}}{Fig.~A}), 
    hence we report its best performance. 
    }
    \label{tab:ablation}
    
\end{table}

\subsection{Ablations} \label{sec:exp_abl}
We conduct thorough ablations on RealEstate10K to analyze \method.
Results are shown in~\cref{tab:ablation} and~\cref{fig:re10k_ablation}, and we discuss them in detail next. Our full model without ``depth refinement'' (Sec.~\ref{sec:mvdepth}) is regarded as the ``base''.

\boldstart{Importance of cost volume.}
The cost volume serves as a cornerstone to the success of \method,
which plays the most important role in our encoder to provide better geometry quality.
To measure our cost volume scheme's importance,
we compare \method to its variant (w/o cost volume) in~\cref{tab:ablation}.
When removing it from the ``base'' model, the quantitative results drop significantly: it decreases the PSNR by more than 3dB, and increases LPIPS by 0.064 (nearly \emph{50\% relative degradation}).
This deterioration is more obviously evident in the rendered visual result in~\cref{fig:re10k_ablation}.
The variant ``w/o cost volume'' exhibits a direct overlay of the two input views,
indicating that the 3D Gaussian parameters extracted from the two input views fail to align within the same 3D space.

\boldstart{Ablations on cross-view attention.}
The cross-view matching is significant in learning multi-view geometry.
The feature extraction backbone in our \method is assembled with cross-view attention to enhance the feature expressiveness by fusing information between input views.
We investigate it by removing the cross-view attention in our transformer.
The quantitative results in~\cref{tab:ablation} (``w/o cross-attn'') show a performance drop of 1dB PSNR,
and the visual results in~\cref{fig:re10k_ablation} (``w/o cross-attn'') showcase higher error intensity.
This highlights the necessity for information flow between views.
Besides, we also observe that this variant ``w/o cross-attn'' suffers from over-fitting (details are shown in the supplementary material \iftoggle{withSupp}{\cref{fig:supp_val_logs}}{Fig.~A}), 
further confirming that this component is critical for model robustness. Due to the over-fitting issue, we report this variant specifically with its best performance instead of the final over-fitted one.

\boldstart{Ablations on the cost volume refinement U-Net.} 
The initial cost volume might be less effective in challenging regions, thus we propose to use a U-Net for refinement. To investigate its importance, we performed a study (``w/o U-Net'') that removes the U-Net architecture. 
\cref{fig:re10k_ablation} reveals that, for the middle regions, the variant ``w/o U-Net'' render as well as the ``base'' model, where content is present in both input views;
but its left and right parts show obvious degraded quality in this variant compared with the ``base'' model,
for which content is only present in one of the inputs.
This is because our cost volume \emph{cannot} find any matches in these regions, leading to poorer geometry cues.
In such a scenario, the U-Net refinement is important for mapping high-frequency details from input views to the Gaussian representation,
resulting in an overall improvement of $\sim 0.7$ dB PSNR as reported in~\cref{tab:ablation}.

\boldstart{Ablations on depth refinement.} 
Additional depth refinement helps improve the depth quality, essentially leading to better visual quality.
As in Tab.~\ref{tab:ablation}, ``base + refine'' achieves the best outcome, hence it is used as our final model.

\boldstart{More ablations.} We further demonstrate in the supplementary material \iftoggle{withSupp}{\cref{sec:app_experiment}}{Appendix~A} 
that our cost volume based design can also greatly enhance pixelSplat~\cite{charatan2023pixelsplat}. Swin Transformer is more suitable for our \method than Epipolar Transformer~\cite{he2020epipolar}. Our \method can benefit from predicting more Gaussian primitives. And our \method is superior to existing methods even when training from completely random initialization for the entire model. 

\section{Conclusion}
\label{sec:conclusion}

We present \method, an efficient feed-forward 3D Gaussian Splatting model that is trained using sparse multi-view images.
The key to our success is that we construct a cost volume to exploit multi-view correspondence information for better geometry learning, which differs from existing approaches that resort to data-driven design. With a well-customized encoder tailored for 3D Gaussians primitives prediction, our \method sets new state-of-the-art in two large-scale scene-level reconstruction benchmarks.
Compared to the latest state-of-the-art method pixelSplat, our \method uses $10\times $ fewer parameters and infers more than $2\times$ faster while providing higher appearance and geometry quality as well as better cross-dataset generalization.

\boldstart{Limitations and Discussions.}
Our model might produce unreliable results for reflective surfaces like glasses and windows, which are currently open challenges for existing methods (detailed in the supplementary material \iftoggle{withSupp}{\cref{sec:app_experiment}}{Appendix~A}).
Besides, our model is currently trained on the RealEstate10K dataset, where its diversity is not sufficient enough to generalize robustly to in-the-wild real-world scenarios despite its large scale.
It would be an interesting direction to explore our model's scalability to larger and more diverse training datasets (\eg, by mixing several existing scene-level datasets) in the future.

\boldstart{Acknowledgements} This research is supported by the Monash FIT Start-up Grant. Dr. Chuanxia Zheng is supported by EPSRC SYN3D EP/Z001811/1.

\bibliographystyle{splncs04}
\bibliography{main}

\clearpage

\appendix
\setcounter{table}{0}
\setcounter{figure}{0}
\renewcommand{\thetable}{\Alph{table}}
\renewcommand{\thefigure}{\Alph{figure}}

\section{More Experimental Analysis} \label{sec:app_experiment}

All experiments in this section follow the same settings as in \cref{sec:exp_abl} unless otherwise specified, which are trained on RealEstate10K~\cite{DBLP:journals/tog/ZhouTFFS18} and reported by averaging over the full test set.

\boldstart{Using cost volume in pixelSplat.} In the main paper, we have demonstrated the importance of our cost volume design for learning feed-forward Gaussian models. We note that such a concept is general and is not specifically designed for a specific architecture. To verify this, we replace pixelSplat's probability density-based depth prediction module with our cost volume-based approach while keeping other components intact. The results shown in \cref{tab:supp_costvolume} again demonstrate the importance of the cost volume by significantly outperforming the original pixelSplat, indicating the general applicability of our proposed method.

\begin{table}[h!]
    \begin{center}
    \begin{tabular}{lccccc}
    \toprule
    Setup  & PSNR$\uparrow$ & SSIM$\uparrow$ & LPIPS$\downarrow$  \\
    \midrule

    pixelSplat (w/ probability density depth)~\cite{charatan2023pixelsplat} & 25.89 & 0.858 & 0.142 \\
    pixelSplat (w/ our \method cost volume depth)  & \textbf{26.63} & \textbf{0.875} & \textbf{0.122} \\
    \bottomrule
    \end{tabular}
    \end{center}
    \caption{\textbf{Using cost volume in pixelSplat}. Our cost volume-based depth prediction approach can also be used in the pixelSplat~\cite{charatan2023pixelsplat} model by replacing its probability density-based depth branch and its performance can be significantly boosted, which demonstrates the general applicability of our method.
    }
    \label{tab:supp_costvolume}
    \vspace{-0.3in}
\end{table}

\begin{table}[h!]
    \begin{center}
    \begin{tabular}{lccccc}
    \toprule
    Setup & Time (s) & PSNR$\uparrow$ & SSIM$\uparrow$ & LPIPS$\downarrow$  \\
    \midrule
    MVSplat (w/ Epipolar Transformer~\cite{charatan2023pixelsplat})  & 0.055 & 26.09 & 0.865 & 0.133 \\
    MVSplat (w/ Swin Transformer) & \textbf{0.038} & 26.12 & 0.864 & 0.133 \\
    \bottomrule
    \end{tabular}
    \end{center}
    \caption{\textbf{Comparisons of the backbone Transformer}. 
    Our Swin Transformer~\cite{liu2021swin}-based architecture is more efficient than the Epipolar Transformer counterpart in pixelSplat~\cite{charatan2023pixelsplat} since the expensive epipolar sampling process is avoided. Besides, there is no clear difference observed in their rendering qualities.
    }
    \label{tab:supp_epitrans}
    \vspace{-0.2in}
\end{table}

\boldstart{Ablations on the backbone Transformer.} Differing from  pixelSplat~\cite{charatan2023pixelsplat} and GPNR~\cite{suhail2022generalizable} that are based on the Epipolar Transformer, we adopt Swin Transformer~\cite{liu2021swin} in our backbone (\ie, Multi-view feature extraction as in \cref{sec:mvdepth}). To compare the Transformer architectures, we conduct ablation experiment by replacing our Swin Transformer with the Epipolar Transformer in \cref{tab:supp_epitrans}. Since the Swin Transformer does not need to sample points on the epipolar line, where the sampling process is computationally expensive, our model is more efficient than the Epipolar Transformer counterpart. Besides, there is no clear difference observed in their rendering qualities, demonstrating the superiority of our Swin Transformer-based design.

\begin{table}[t]
    \begin{center}
    \begin{tabular}{lccccc}
    \toprule
    Setup  & Render Time (s)  & PSNR$\uparrow$ & SSIM$\uparrow$ & LPIPS$\downarrow$  \\
    \midrule

    MVSplat (1 Gaussian per pixel) & 0.0015 & 26.39 & 0.869 & 0.128 \\
    MVSplat (3 Gaussians per pixel) & 0.0023 & 26.54 & 0.872 & 0.127 \\
    \bottomrule
    \end{tabular}
    \end{center}
    \caption{\textbf{Comparisons of Gaussians' number per pixel}. Increasing the number of Gaussians improves the performance but slows down the rendering speed.
    }
    \label{tab:supp_gpp}
\end{table}

\boldstart{Ablations on the Gaussian numbers per pixel.} 
Unlike pixelSplat that predicts three Gaussians per image pixel, our \method by default only predicts one per pixel. However, \method can also benefit from increasing the number of Gaussians. As reported in \cref{tab:supp_gpp}, our default model can be further boosted by predicting more Gaussians, but it also impacts the rendering speed. We choose to predict one Gaussian per pixel to balance performance and rendering speed.

\begin{table}[t]
    \begin{center}
    \begin{tabular}{lccccc}
    \toprule
    Method  & PSNR$\uparrow$ & SSIM$\uparrow$ & LPIPS$\downarrow$  \\
    \midrule

    pixelSplat w/ DINO init (300K) ~\cite{charatan2023pixelsplat}              & 25.89 & 0.858 & 0.142 \\

    \midrule

    MVSplat w/ UniMatch init (300K)               & 26.39 & 0.869 & 0.128 \\
    MVSplat w/ random init (300K) & 26.01 &  0.863 & 0.133  \\
    MVSplat w/ random init (450K) & 26.29 & 0.868 & 0.128 \\
    \bottomrule
    \end{tabular}
    \end{center}
    \caption{\textbf{Comparisons of backbone initialization}. By default we train our models for 300K iterations with the publicly available UniMatch pretrained weights as initialization. However, our model can also be trained with random initialization (``w/ random init'') and still outperforms previous state-of-the-art method pixelSplat. The random initialized model needs more training iterations (450K) to reach the performance of the UniMatch initialized model.
    }
    \label{tab:supp_sota_variants}
\end{table}

\boldstart{Ablations on backbone initialization.} In our implementations, we initialize our backbone (\ie, Multi-view feature extraction as in \cref{sec:mvdepth}) with the publicly available UniMatch~\cite{xu2023unifying} pretrained weight\footnote{\url{https://github.com/autonomousvision/unimatch}}, where its training data has no overlap with any datasets used in our experiments.
However, our model can also be trained with random initialization (``w/ random init'') and still outperforms previous state-of-the-art method pixelSplat~\cite{charatan2023pixelsplat}, whose backbone is initialized with the weights pretrained on ImageNet with a DINO objective. To compensate for the lack of a proper initialization, our random initialized model needs more training iterations (450K) to reach the performance of the UniMatch initialized model.
The results further demonstrate our model's high efficiency and effectiveness, where strong performance can still be obtained without relying on pretraining on large-scale datasets.

\begin{figure}[t!]
    \centering
    \includegraphics[width=\textwidth]{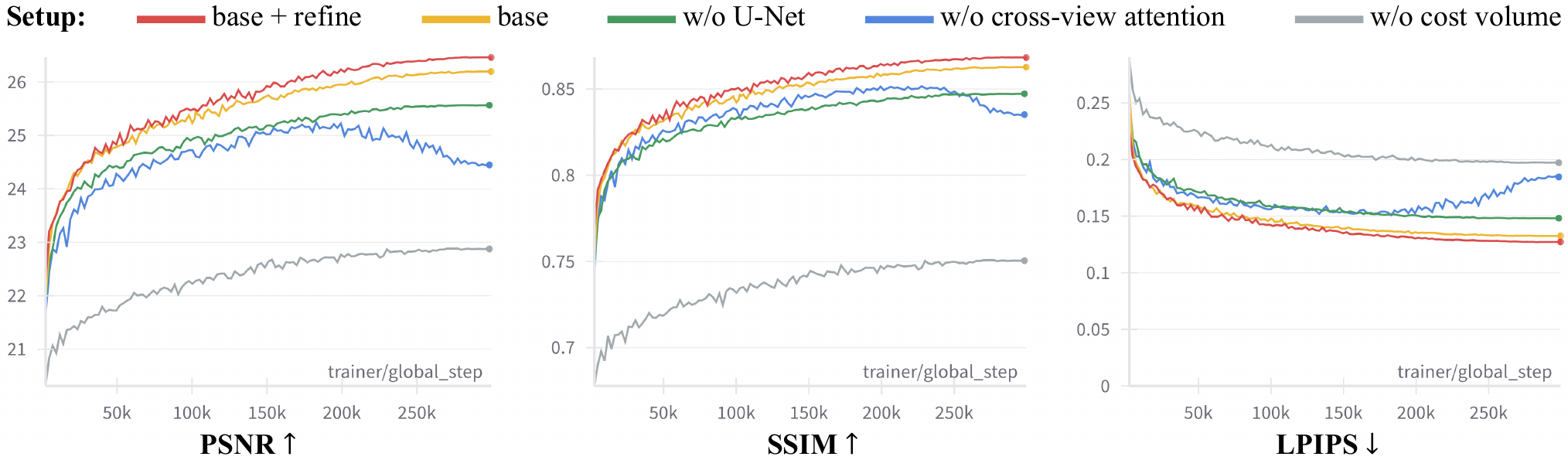}
    \caption{\textbf{Validation curves of the ablations}. 
    The setup of each model is illustrated on the top, which refers to the same one as in \cref{tab:ablation} of the main paper. The cost volume plays a fundamental role in our full model, and interestingly, the model without cross-view attention suffers from over-fitting after certain training iterations.
    }
    \label{fig:supp_val_logs}
\end{figure}

\boldstart{Validation curves of the ablations}. To better perceive the performance difference of different ablations, we provide the validation curves throughout the whole training phase in \cref{fig:supp_val_logs}). 
The cost volume plays a fundamental role in our full model, and interestingly, the model without cross-view attention suffers from overfitting after certain training iterations.

\begin{figure}[t!]
    \centering
    \includegraphics[width=\textwidth]{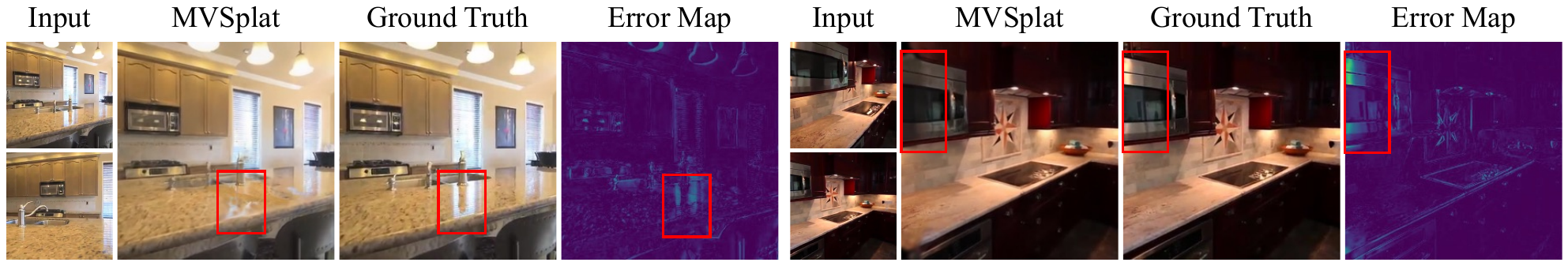}
    \caption{\textbf{Failure cases}. Our \method might be less effective on the non-Lambertian and reflective surfaces. 
    }
    \label{fig:supp_limitation}
\end{figure}

\boldstart{Limitation.} Our \method might be less effective on non-Lambertian and reflective surfaces, as shown in \cref{fig:supp_limitation}.
Integrating the rendering with additional BRDF properties and training the model with more diverse datasets might be helpful for addressing this issue in the future.

\begin{figure}[t!]
    \centering
    \includegraphics[width=\textwidth]{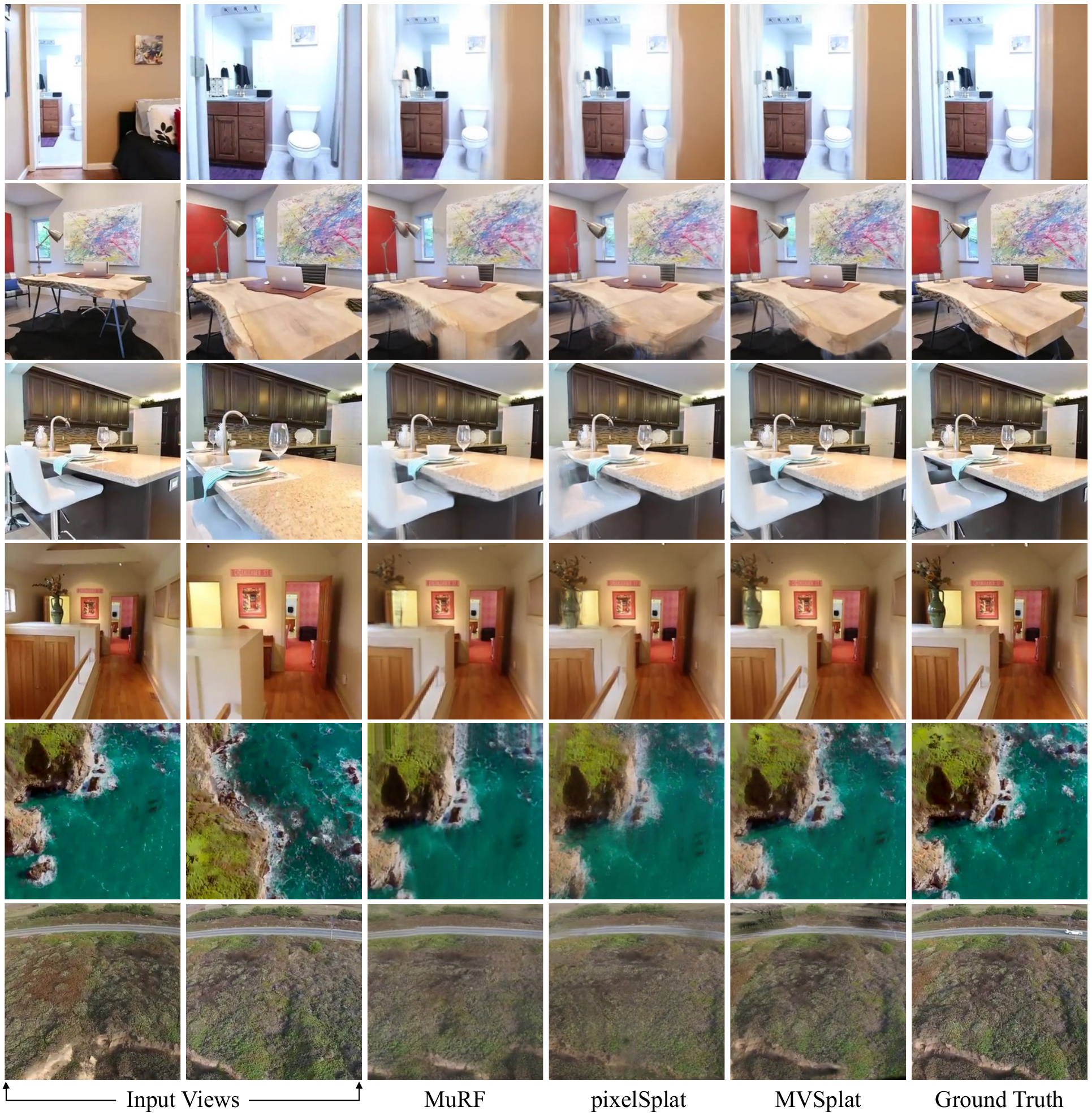}
    \caption{\textbf{More comparisons with the state of the art}. These are the extended visual results of \cref{fig:sota_comparisons}. Scenes of the first four rows come from the RealEstate10K, whilst scenes of the last two rows come from ACID. Our \method performs the best in all cases.
    }
    \label{fig:supp_sota}
\end{figure}

\boldstart{Potential negative societal impacts.} Our model may produce unreliable outcomes, particularly when applied to complex real-world scenes. Therefore, it is imperative to exercise caution when implementing our model in safety-critical situations, \eg, when augmenting data to train models for autonomous vehicles with synthetic data rendered from our model.

\section{More Visual Comparisons} \label{sec:app_visual}

\begin{figure}[t!]
    \centering
    \includegraphics[width=\textwidth]{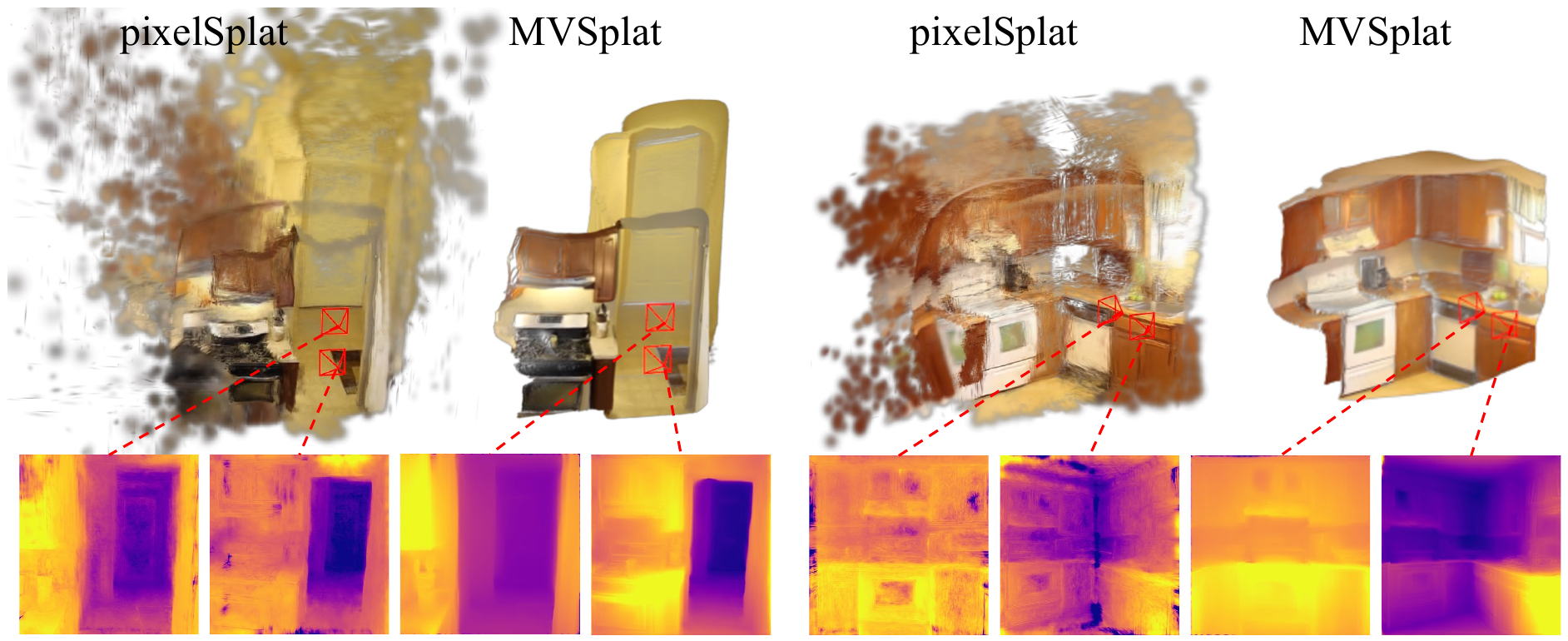}
    \caption{\textbf{More comparisons of 3D Gaussians (top) and depth maps (bottom)}. These are the extended visual results of \cref{fig:point_cloud}. Extra depth-regularized fine-tuning is \emph{not} applied to either model. 3D Gaussians and depth maps predicted by our \method are both of higher quality than those predicted by pixelSplat.
    }
    \label{fig:supp_point_cloud}
\end{figure}

\begin{figure}[t!]
    \centering
    \includegraphics[width=\textwidth]{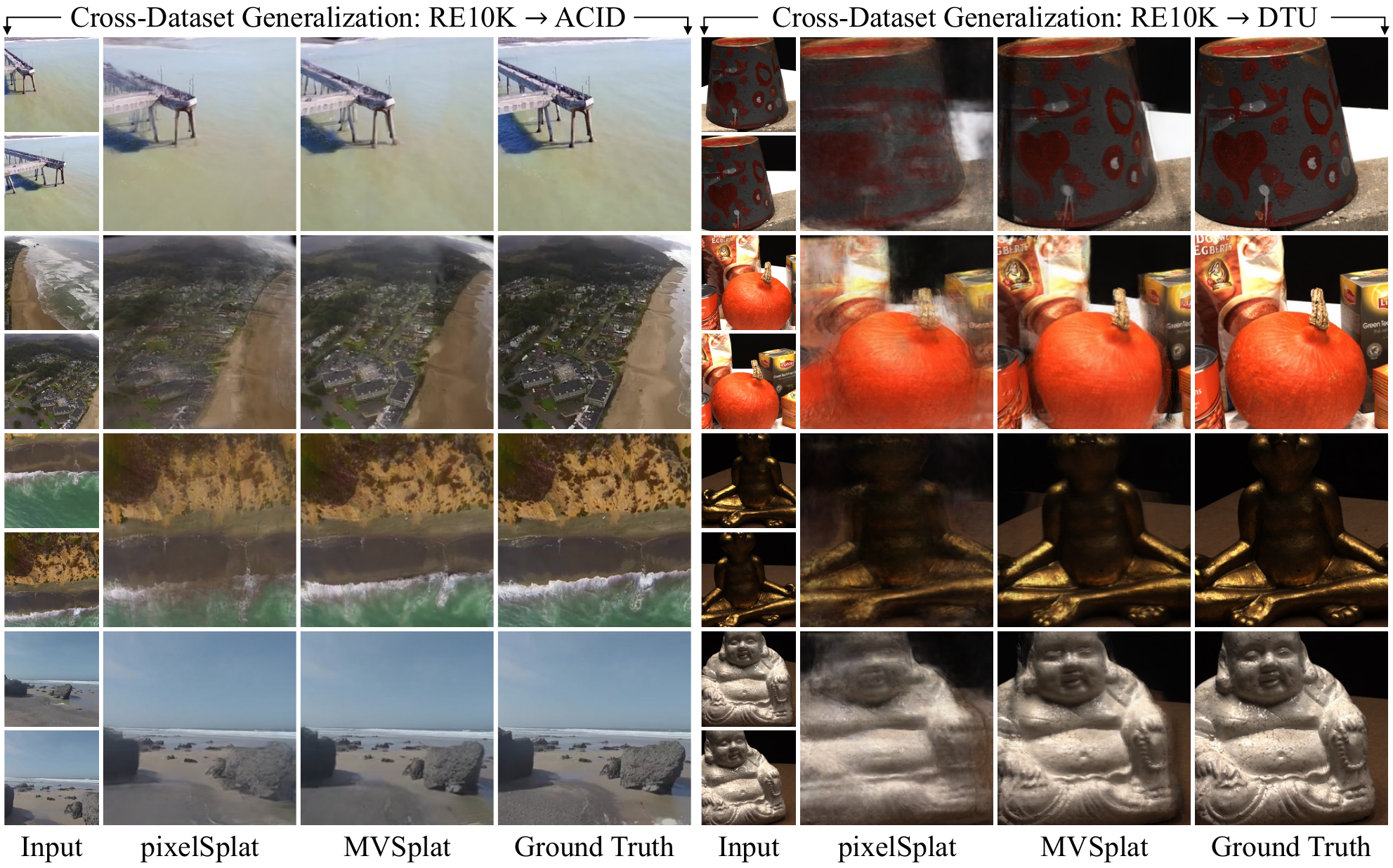}
    \caption{\textbf{More comparisons of cross-dataset generalization}. These are the extended visual results of \cref{fig:re10k_generalization}. By training on indoor scenes (RealEstate10K), our \method generalizes much better than pixelSplat to outdoor scenes (ACID) and object-centric scenes (DTU), showing the superior of our cost volume-based architecture.
    }
    \label{fig:supp_generalization}
\end{figure}

In this section, we provide more qualitative comparisons of our \method with state-of-the-art methods on the RealEstate10K and ACID in \cref{fig:supp_sota}. We also provide more comparisons with our main comparison model pixelSplat~\cite{charatan2023pixelsplat} regarding geometry reconstruction (see \cref{fig:supp_point_cloud}) and cross-dataset generalizations (see \cref{fig:supp_generalization}). Besides, readers are referred to the project page for the rendered videos and 3D Gaussians models (provided in ``.ply'' format).

\section{More Implementation Details} \label{sec:app_implementation}

\boldstart{Network architectures.} Our shallow ResNet-like CNN is composed of 6 residual blocks~\cite{he2016deep}, In the last 4 blocks, the feature is down-sampled to half after every 2 consecutive blocks by setting the convolution stride to 2, resulting in overall $4 \times$ down sampling. The following Transformer consists of 6 stacked Transformer blocks, each of which contains one self-attention layer followed by one cross-attention layer. Swin Transformer's~\cite{liu2021swin} local window attention is used and the features are split into $2 \times 2$ in all our experiments. For the cost volume refinement, we adopt the 2D U-Net implementations from~\cite{rombach2022high}. We keep the channel dimension unchanged throughout the U-Net as 128, and apply 2 times of $2 \times$ down-sampling, with an additional self-attention layer at the $4 \times$ down-sampled level. Inspired by existing multi-view based models~\cite{shi2023mvdream,tang2024lgm}, we also flatten the features before applying self-attention, allowing information to propagate among different cost volumes. The following depth refinement U-Net also shares a similar configuration, except that we apply 4 times of $2 \times$ down-sampling and add attentions at the $16 \times$ down-sampled level. %

\boldstart{More training details.} As aforementioned, we initialize the backbone of \method in all experiments with the UniMatch~\cite{xu2023unifying} pretrained weight. Our default model is trained on a single A100 GPU. The batch size is set to 14, where each batch contains one training scene, including two input views and four target views. Similar to pixelSplat~\cite{charatan2023pixelsplat}, the frame distance between two input views is gradually increased as the training progressed. For both RealEstate10K and ACID, we empirically set the near and far depth plane to 1 and 100, respectively, while for DTU, we set them to 2.125 and 4.525 as provided by the dataset.

\end{document}